\documentclass[conference]{IEEEtran}
\IEEEoverridecommandlockouts

\IEEEpubid{\begin{minipage}{\textwidth}\ \\[12pt]
		 \copyright 2021 IEEE. 2021 IEEE Radar Conference (RadarConf 2021).
\end{minipage}} 

\usepackage{cite}
\usepackage{amsmath,amssymb,amsfonts}
\usepackage{algorithmic}
\usepackage{graphicx}
\usepackage{textcomp}
\usepackage{xcolor}

\usepackage{graphicx} 
\usepackage{subcaption}
\usepackage{tabularx}
\usepackage{booktabs}	
\usepackage{array}
\usepackage{xfrac}	

\def\BibTeX{{\rm B\kern-.05em{\sc i\kern-.025em b}\kern-.08em
    T\kern-.1667em\lower.7ex\hbox{E}\kern-.125emX}}

\makeatletter
\def\endthebibliography{%
	\def\@noitemerr{\@latex@warning{Empty `thebibliography' environment}}%
	\endlist
}
\makeatother

\graphicspath{{./figures/}}

\begin{document}

\title{Deep Evaluation Metric: \\Learning to Evaluate Simulated Radar Point Clouds for Virtual Testing of Autonomous Driving\\
}

\author{
	\IEEEauthorblockN{Anthony Ngo\IEEEauthorrefmark{1}, Max Paul Bauer\IEEEauthorrefmark{1} and Michael Resch\IEEEauthorrefmark{2}}
	\IEEEauthorblockA{\IEEEauthorrefmark{1}Robert Bosch GmbH, Automated Driving, Stuttgart, Germany\\
		\IEEEauthorrefmark{2}University of Stuttgart, High Performance Computing Center, Stuttgart, Germany\\
		Email: Anthony.Ngo@de.bosch.com}
} 

\maketitle

\begin{abstract}
	The usage of environment sensor models for virtual testing is a promising approach to reduce the testing effort of autonomous driving. However, in order to deduce any statements regarding the performance of an autonomous driving function based on simulation, the sensor model has to be validated to determine the discrepancy between the synthetic and real sensor data.
Since a certain degree of divergence can be assumed to exist, the sufficient level of fidelity must be determined, which poses a major challenge. In particular, a method for quantifying the fidelity of a sensor model does not exist and the problem of defining an appropriate metric remains.
In this work, we train a neural network to distinguish real and simulated radar sensor data with the purpose of learning the latent features of real radar point clouds. Furthermore, we propose the classifier's confidence score for the `real radar point cloud' class as a metric to determine the degree of fidelity of synthetically generated radar data.
The presented approach is evaluated and it can be demonstrated that the proposed deep evaluation metric outperforms conventional metrics in terms of its capability to identify characteristic differences between real and simulated radar data.
\end{abstract}


\section{Introduction} 
\label{sec: introduction}
Autonomous driving has the potential to improve road safety and optimize traffic flow while being currently one of the main challenges in the automotive industry \cite{schumann_scene_2020}.
The robust perception and comprehension of the environment of a self-driving vehicle is a substantial topic in this field. Automotive radar is widely employed within modern advanced driver assistance systems and is a key technology for autonomous driving \cite{dickmann_automotive_2016}. A radar sensor uses electromagnetic waves to determine the existence and location of reflecting objects by relying on the strength of received waves \cite{gamba_radar_2020}. By exploiting the Doppler effect, a radar can directly measure the radial velocity of an object and it works reliably even in adverse weather conditions \cite{skolnik_radar_2008}. In this way, targets can not only be detected, but static and dynamic objects can be further distinguished and tracked over time, which allows an understanding of the surrounding scene to be built up \cite{schumann_scene_2020}.

In addition to the functional development of the perception functions for autonomous driving, the validation of such a system poses a major difficulty \cite{junietz_evaluation_2018}. As a statistical validation of safety based on field testing is not economically feasible, novel approaches are needed \cite{stellet_validation_2020}. The usage of sensor data generated in a virtual environment is a promising approach to enable efficient testing of autonomous driving functions \cite{boede_efficient_2018}.

However, in order to allow any implications about the real system based on virtual testing the employed sensor models have to be validated \cite{rosenberger_towards_2019}. It is therefore essential to determine the requirements a radar simulation must fulfill. 
Although many approaches to simulate a radar sensor have been reported in the literature, there exists no generally accepted method to evaluate simulated radar data \cite{holder_measurements_2018}. A method for quantifying the fidelity of a sensor model does not yet exist and the problem of defining an appropriate metric remains, since a qualitative evaluation relying on a visual matching does not scale. Thus, a method that provides an objective and quantitative evaluation of synthetically generated radar data is needed.

\begin{figure}[thpb]
	\centering
	\setlength{\fboxrule}{0pt}
	\framebox{\parbox{\linewidth}{
			\includegraphics[width=\linewidth] {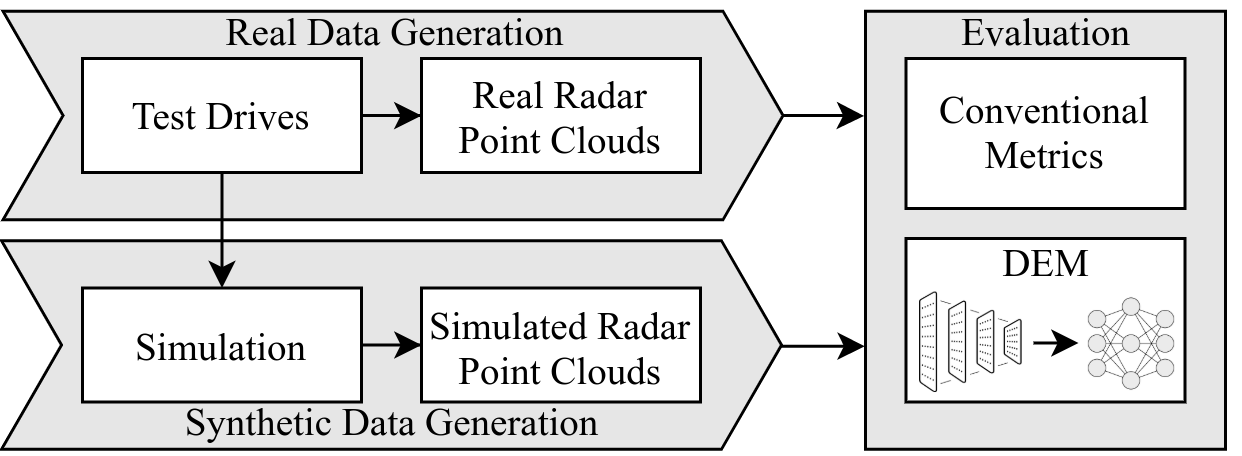}
	}}
	\caption{Paper overview with the proposed Deep Evaluation Metric (DEM).}
	\label{fig: method overview}
\end{figure}

Therefore, a machine learning-based approach to evaluate a radar simulation is presented in this work (see Fig. \ref{fig: method overview}). We train a neural network (PointNet++ \cite{qi_pointnet_2017}) to classify real and simulated radar sensor data with the purpose of learning the characteristic features of real radar point clouds. Furthermore, we propose the classifier's confidence score of the `\textit{real radar point cloud}' class as a metric to determine the degree of fidelity of synthetically generated radar data.

Our main contributions are:
\begin{itemize}
	\item a study on the state-of-the-art evaluation approaches of a radar simulation
	\item a conventional evaluation with existing metrics
	\item a proposal of a novel machine-learning based evaluation metric: Deep Evaluation Metric (DEM)
	\item an evaluation of both approaches by analyzing the metrics on a real-world data set
\end{itemize}

The rest of this paper is organized as follows: Section \ref{sec: related work} gives a brief overview of the existing approaches to evaluate a radar simulation. The proposed method is presented at length in Section \ref{sec: method}. Section \ref{sec: experiments and results} elaborates on the conducted experiments and discusses the effectiveness of the introduced method. Finally, Section \ref{sec: conclusion} concludes this paper with a concise outlook on further research.

\section{Related Work}
\label{sec: related work}
The following provides an overview of existing approaches to evaluate synthetically generated radar data, sorted by the degree of abstraction of the radar model output: raw data level, detection level and perception level.

\subsection{Raw Data Level Evaluation}
A radar simulation is often composed of individual sub-modules, which approximate different components and physical effects of the electromagnetic wave propagation and the radar signal processing \cite{bernsteiner_radar_2015}. In this regard, the abstract raw data level represents any level before a radar detection is generated and is more complicated to evaluate due to the stochastic nature of a radar sensor.
The characteristics of a reflecting object at a certain range is represented by the radar cross section (RCS) \cite{gamba_radar_2020} and different modeling approaches can be found in the literature. The evaluation of these approaches varies from simple qualitative observations to an assessment using defined metrics. Besides the qualitative evaluation, Owaki and Machida \cite{owaki_hybrid_2019} use a correlation coefficient between ground truth and estimated RCS. Furthermore, Deep et al. \cite{deep_radar_2020} introduce the following metrics to also analyze the spectrum of the received radar signals: Normalized mean square error (NMSE), the structural similarity index (SSIM), the normalized cross-correlation (NCC) and mutual information (MI).
In contrast to the comparison with real measurement data, there are also approaches that analyze the quality of synthetically generated radar data only in simulation \cite{chipengo_high_2020}.
This evaluation approach has the downside that only simple scenarios and the basic functionality can be tested where each phenomenon and result can be specifically reasoned and described. 

\subsection{Detection Level Evaluation}
The purpose of the target detection is to distinguish genuine object reflections from noise and clutter \cite{skolnik_radar_2008}. In this work, the detection level refers to the interface after a reflection passed the detection threshold, resulting in the radar point cloud. To the best knowledge of the authors,  the evaluation on this level is relatively unexplored and there mainly exists qualitative evaluations \cite{martowicz_uncertainty_2019}. For lidar point clouds, which are comparable to the radar detection interface, various methods can be found in the literature. These approaches range from purely visual comparisons \cite{hirsenkorn_learning_2017} to distance based metrics \cite{browning_3d_2012} and occupancy grids \cite{schaermann_validation_2017}. Nevertheless, the question arises whether these metrics can be used to evaluate synthetic radar point clouds, considering that radar data is more sparse and stochastic in nature compared to lidar data.

\subsection{Perception Level Evaluation}
Up to this point, the detections are neither clustered, nor interpreted as objects. At the perception level the detections are further processed to build up an understanding of the surrounding scene. Holder et al. \cite{holder_how_2020} present a method to evaluate a radar simulation by feeding simulated data into an algorithm developed and parameterized on real radar data in order to qualitatively investigate the strength and weaknesses of the sensor simulation. Bernsteiner et al. \cite{bernsteiner_radar_2015} compare a tracking algorithm result between simulated and real data qualitatively. Moreover, Jasinski \cite{jasinski_generic_2019} proposes a similar approach by evaluating a radar simulation indirectly with a tracking algorithm. Although the author suggests a quantitative concept with the intersection over union (IoU) as a metric, the results are not provided. In general, the evaluation of perception algorithms are more investigated and matured in comparison with the two preceding levels. However, it needs to be further researched whether these metrics are suitable to evaluate synthetically generated sensor data.

\section{Method}
\label{sec: method}
The method introduced in this section focuses on the enhancement of existing approaches at the detection level by incorporating a quantitative evaluation without the need for handcrafted metrics. Therefore, we propose a machine learning-based approach and compare the results with conventional methods to evaluate synthetically generated radar point clouds. The method consists of the following four main steps (see Fig. \ref{fig: method overview}): real and synthetic data generation, conventional metrics as well as the proposed deep evaluation metric.

\subsection{Real Data Generation}
The first step comprises the generation of real radar data as a reference for evaluation. A comparison with real radar data is essential in order to permit any prediction about the real system behavior from virtual testing.
In this work, the test drives are conducted on a testing site with the ego vehicle and one target vehicle.
A differential global positioning system (DGPS) with an inertial measurement unit is used for a precise acquisition of the position, orientation and velocity of the vehicles. A high degree of accuracy is crucial, because the resulting ground truth data serves as the basis to reproduce the same scenarios in a virtual environment. The radar sensor data is recorded with various scenarios ranging from stationary tests to overtaking maneuvers.

\subsection{Synthetic Data Generation}
\begin{figure}[thpb]
	\centering
	\setlength{\fboxrule}{0pt}
	\framebox{\parbox{\linewidth}{
			\includegraphics[width=\linewidth] {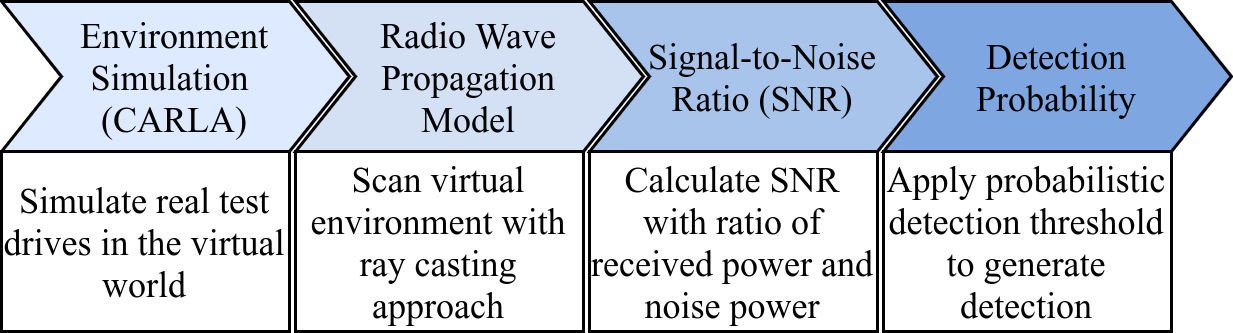}
	}}
	\caption{Radar simulation processing pipeline.}
	\label{fig: radar simulation pipeline}
\end{figure}

The generation of synthetic data is mainly divided in two steps: the simulation of real test drives based on the recorded ground truth data, and the generation of a virtual scene of the environment from the sensor point of view, resulting in the simulated radar point cloud.
The process of the latter is depicted in Fig. \ref{fig: radar simulation pipeline} and is briefly explained in the following. The implementation details of the underlying submodules and formulas used are thoroughly explained in \cite{ngo_sensitivity_2020}.

\subsubsection{Environment Simulation}
The open-source simulator CARLA \cite{dosovitskiy_carla_2017} is used to implement the outlined method and the testing site is virtually reproduced in the simulation. This virtual environment is perceived by a radar sensor model in the simulation to generate the synthetic radar point clouds.

\subsubsection{Radio Wave Propagation Model}
The employed radar simulation approximates the propagation of electromagnetic waves with a ray casting approach based on the geometric optics diffraction theory, in which radio waves are modeled as a bundle of rays \cite{keller_geometrical_1962}. Each beam hitting an object within the sensor's field of view returns a reflection.

\subsubsection{Signal-to-Noise Ratio}
Subsequently, the generated reflections are further processed by calculating the signal-to-noise ratio (SNR) at each location. The SNR describes in general the performance of a radar sensor and can be expressed by the ratio between the received signal power and the noise power \cite{skolnik_radar_2008}.

\subsubsection{Detection Probability}
In order to generate detections based on the calculated SNR a detection threshold is applied in the final step. This way, target returns can be distinguished from the prevailing noise and clutter \cite{skolnik_radar_2008}. We furthermore incorporate detection probabilities with the purpose of approximating the stochastic behavior of noise.

\subsection{Conventional Metrics} \label{sec: method conventional evaluation}
In this section, the conventional evaluation metrics are introduced. We implemented two different metrics to analyze the characteristics of a radar point cloud.

 Since each radar detection is defined in this work by its two-dimensional location and the Doppler velocity, both components are compared to evaluate the difference between the simulated and the real radar point cloud. In that respect, we use the normalized sum of the smallest Euclidean distance from every point in the real point cloud $X = (x_1 , ..., x_M )$ to the simulated point cloud $Y = ( y_1, ..., y_N )$, where $x_m, y_n \in \mathbb{R}^{3}$ are three-dimensional points. This point cloud to point cloud distance is first introduced by Browning et al. \cite{browning_3d_2012} and is defined as: 
\begin{equation} \label{eq: dpp}
D'_{pp}(X,Y) := \frac{1}{M} \sum_{m=1}^{M} \min_{1\leq n\leq N} ||x_m - y_n||.
\end{equation}
This approach has the benefit that the difference in values of each point as well as the difference in the number of points between both point clouds are considered. Moreover, it is divided by the respective number of points for normalization. Since $D'_{pp}$ is a non-symmetrical distance metric, the worst-case is assumed: 
\begin{equation} \label{eq: dpp max}
D_{pp}(X,Y) := \mathrm{max}(D'_{pp}(X,Y), D'_{pp}(Y,X)).
\end{equation}

For the second metric, we propose the Wasserstein distance also known as the earth mover's distance (EMD) to compare the point distributions of different radar point clouds. Given that the EMD is based on the Kantorovich-Rubinstein theorem \cite{kantorovich_space_1958} concerning the optimal transportation problem \cite{hitchcock_distribution_1941}, it measures the disparity between two distributions by the optimal cost of rearranging one distribution into the other:

\begin{equation} \label{eq: emd}
EMD(X,Y) := \frac{\sum_{m=1}^{M} \sum_{n=1}^{N} f_{m,n}d_{m,n}}{\sum_{m=1}^{M} \sum_{n=1}^{N} f_{m,n}}.
\end{equation}
Apart from the three-dimensional point clouds $X$ and $Y$, $m$ and $n$ describe the number of points in the point sets and the solution to the transportation problem between both point cloud distributions is expressed by the optimal flow $f_{m,n}$. In this paper, the Euclidean distance is chosen as the ground distance $d_{m,n}$. Thus, EMD naturally extends the notion of a distance between single points to that of a distance between distributions of points. A detailed derivation of the stated equation can be found in Rubner et al. \cite{rubner_earth_2000}.

\subsection{Deep Evaluation Metric}
The conventional evaluation approach relies on self-defined metrics which evaluates specific characteristics like the spatial distribution between real and simulated point clouds. The problem of selecting the right metric remains, which is tantamount to deciding which characteristics or physical effects are most important to consider.

This section introduces a machine learning-based metric to evaluate the fidelity of synthetically generated sensor data. The objective of the proposed method is to train a neural network to be able to classify real and simulated radar data. In contrast to the conventional evaluation, the intention thereby is to learn the latent features that differentiate real from simulated radar point clouds without having to determine in advance which characteristics to consider specifically. Furthermore, we propose the classifier's predicted confidence score of the '\textit{real radar point cloud}' class as a metric to determine the degree of fidelity of synthetically generated radar data.

In the following, the process of selecting and adjusting a suitable network architecture is presented in addition to the used data set along with the training and testing of the network.

\subsubsection{Network Architecture}
Since the input of most neural networks follow a regular structure like a grid map representation, data such as radar point clouds have to be transformed to a regular format before feeding them into a neural network. Qi et al. provide with PointNet++ \cite{qi_pointnet_2017} a method to overcome this constraint and work directly with point clouds so that no previous mapping is needed. PointNet++ is a hierarchical neural network which is able to learn local features and handle point sets with varying densities. Additionally taken into consideration that Schumann et al. \cite{schumann_semantic_2018} and Danzer et al. \cite{danzer_2d_2019} have shown that this network works well on radar point clouds, the PointNet++ architecture is used for our approach.

\subsubsection{Data set}
For this proof of concept implementation only the radar detections around the target vehicle are considered. Due to the fact that the sensor data was recorded on an empty test site, this is a reasonable simplification. These real test drives are reproduced in simulation to generate the respective synthetic radar data. As a consequence, the resulting data set is quite balanced between real and simulated point clouds. The data set comprises 235 scenarios, corresponding to $1.59\times10^5$ point clouds with $3\times10^6$ radar detections in total. Each detection of a radar point cloud fed into the network contains two spatial coordinates along with the Doppler velocity. The whole data set is randomly split into a training and testing set with a 70/30 ratio.

\subsubsection{Training and Testing}
The architecture is trained from scratch, using both the real radar data and the synthetically generated radar data. Furthermore, the data set is augmented during training to avoid model overfitting. The sensor data is therefore perturbed using random Gaussian noise with zero mean and
standard deviation of 0.1. Random noise is applied to each feature dimension, so that the spatial positions of the detections as well as the Doppler velocities of both real and simulated sensor data are altered.
To ensure a fixed number of input points for each point cloud, sampling is performed, by means of randomly duplicating (oversampling) or drawing (undersampling) up to 10 detections from a point cloud.
The initial learning rate of the model is chosen to be 0.001 and the batch size for training is 32. Training of the model uses the Adam optimizer and is performed for 30 epochs on two NVIDIA GeForce RTX 2080 Ti GPUs.
During testing, the batch size is set to 1 in order to allow a variable number of points to be processed. The network achieves a classification accuracy of 82.14\% within the testing set.

\section{Experiments \& Results}
\label{sec: experiments and results}
First, the experimental setup is presented in this section. To ensure that the network has learned the latent features that distinguish both real and synthetic point clouds and is therefore able to differentiate them, the performance of the trained network is assessed.
Building on this, it is investigated whether the output of the final network layer (the confidence score of the `real radar point cloud' class) can be used as an evaluation metric to indicate the sensor model fidelity. For this reason, the proposed deep evaluation metric is evaluated along with conventional metrics and the effectiveness of both methods is compared and discussed.

\subsection{Experimental Setup and Classification Performance}
\begin{figure}[thpb]
	\centering
	\setlength{\fboxrule}{0pt}
	\framebox{\parbox{\linewidth}{
			\includegraphics[width=\linewidth] {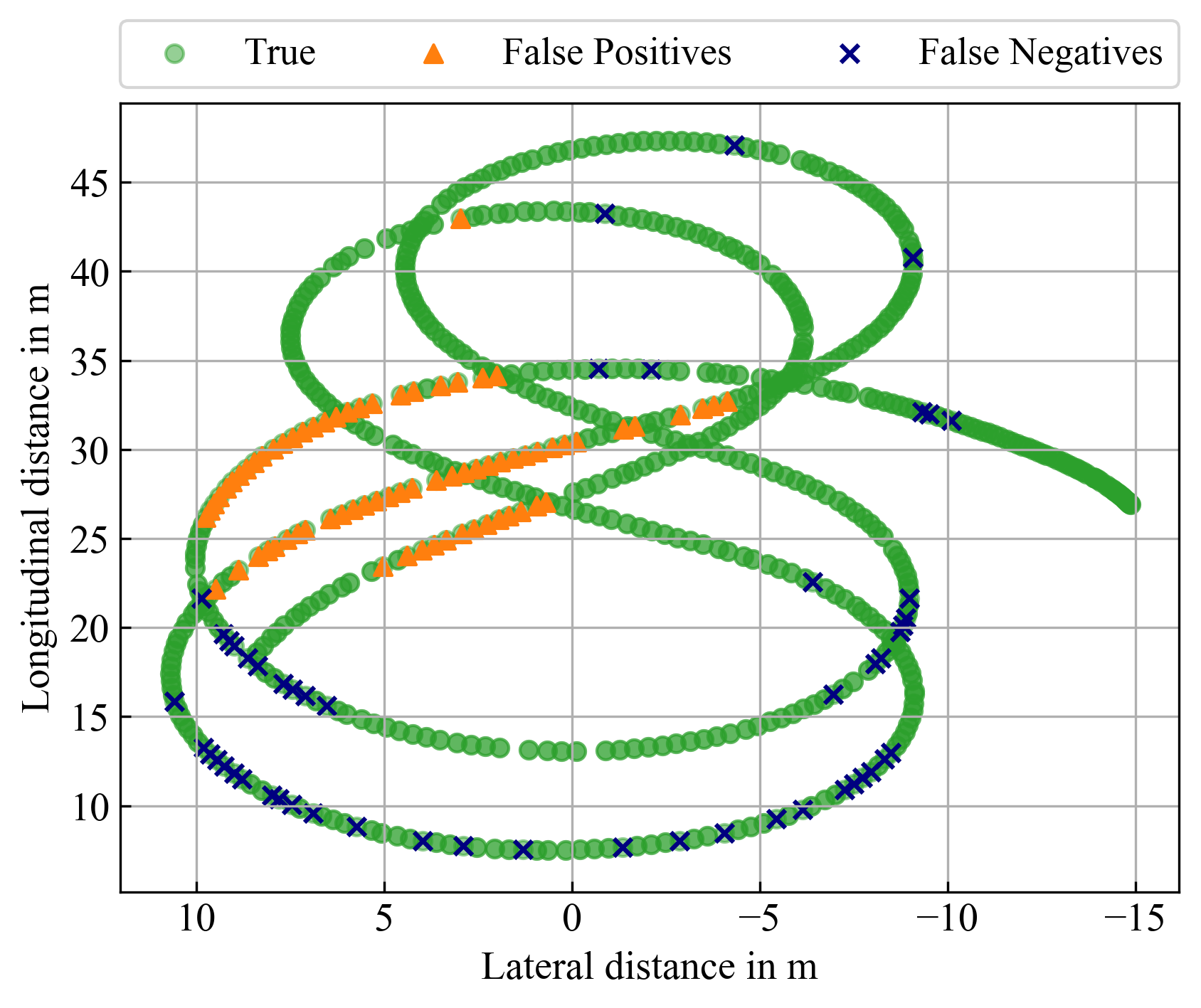}
	}}
	\caption{Classification results on a withheld scenario. The model is fed with real radar data as well as the corresponding simulated radar data. The green detections indicate all correct class predictions. Additionally, the false positives (input: simulated, prediction: real) and false negatives (input: real, prediction: simulated) are depicted.}
	\label{fig: predicted class}
\end{figure}

To ensure comparability, both approaches are evaluated using the same scenario in which a target vehicle drives a path in the shape of an eight in front of the radar sensor, which is static in (0,~0) (see Fig. \ref{fig: predicted class}). Since it can be assumed that the real radar detections change in density and distribution over different positions and orientations of the target vehicle, the objective of this scenario is to analyze whether and to what extent the radar simulation is capable to model this behavior. 

Given that the used radar simulation includes a random component (detection probability) to approximate the stochastic behavior of the real radar data, the evaluation results are subject to random effects. With the purpose of diminishing these effects, the scenario was simulated 100 times and the results are averaged over these runs.

Apart from the driven path, the classification result of the trained network is color coded in Fig. \ref{fig: predicted class}. In addition to the real radar data from the test drive, the present scenario is reproduced in simulation and the resulting synthetic radar point clouds are fed into the network with the intention to examine its capability to distinguish between real and simulated point clouds.
This particular scenario was withheld from the training and testing set in order to guarantee an unbiased performance evaluation.

In the present scenario, the trained model achieves a classification accuracy of 91.99\% with real and 88.59\% with simulated radar point clouds as input. It is evident that most of the misclassifications are located in certain regions for both inputs. The center of the false positives can be observed in a longitudinal distance of around 27 meters and a lateral distance of approximately 5 meters.
This indicates that either the target car was not seen enough in the training before in this zone or that this region exhibits a weakness of the radar sensor model. On the contrary, the majority of the false negatives are found in the near longitudinal distance and are distributed along the lateral axis.

In summary, the network has predicted most of the real and synthetic radar point clouds correctly in this scenario, which is an indication that the network could learn the characteristic features that differentiate the real and simulated radar data.
This allows us to investigate the proposed deep evaluation metric further and compare it with the conventional methods.

\subsection{Results of Evaluation Approaches}
Besides the averaging of the results over all 100 simulation runs, the data are further processed to ensure a valid comparison between the different metric results. Since the resulting range of values can vary widely, a min-max normalization is applied, which consists of rescaling the range of data to [0,~1]. Furthermore, the axes are reversed in such a way that zero expresses the worst (low sensor model fidelity) and one the best possible value (high sensor model fidelity). As a last step of the post processing, the Savitzky-Golay filter \cite{savitzky_smoothing_1964} is applied for the purpose of smoothing the data in order to better visualize and compare the trend of the different results.

\begin{figure}[thpb]
	\centering
	\setlength{\fboxrule}{0pt}
	\framebox{\parbox{\linewidth}{
			\includegraphics[width=\linewidth] {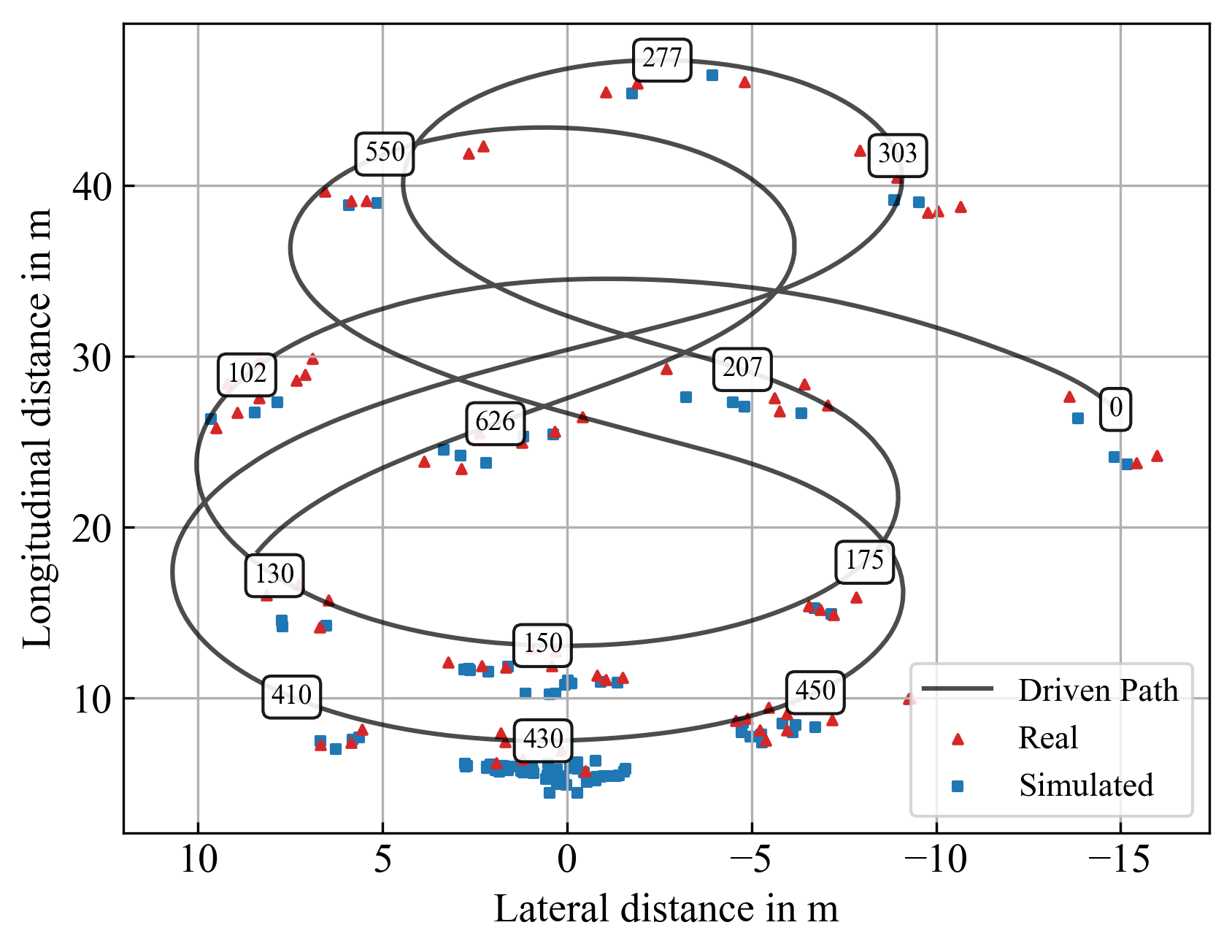}
	}}
	\caption{The real and simulated radar detections and the white boxes indicate the frame number.}
	\label{fig: radar detections}
\end{figure}

In the following, the main differences between the real and simulated radar data are defined, which are identified by a qualitative evaluation based on a visual matching of both sensor data (see Fig. \ref{fig: radar detections}). Based on this, the metrics are then assessed to what extent they can quantifiably reproduce the observed qualitative discrepancies.

Since the presented radar simulation utilizes a ray casting based approach, it can be observed that on the one hand large differences between real and simulated data occur especially at close range due to an increase in the number of simulated detections. On the other hand, the number of points decreases too much with increasing distance compared to the real data. 
An additional effect, which occurs particularly in the closer area, is the formation of an L-shaped point cloud. This is caused by the fact that only the outer shell of the vehicle model is modeled and the aggregated high number of points in the close range allow the edges of the shell to be clearly noticeable. However, this point cloud shape is rather untypical for radar data, because in general there are also detections inside the vehicle.

\begin{figure}[thpb]
	\centering
	\setlength{\fboxrule}{0pt}
	\framebox{\parbox{\linewidth}{
			\includegraphics[width=\linewidth] {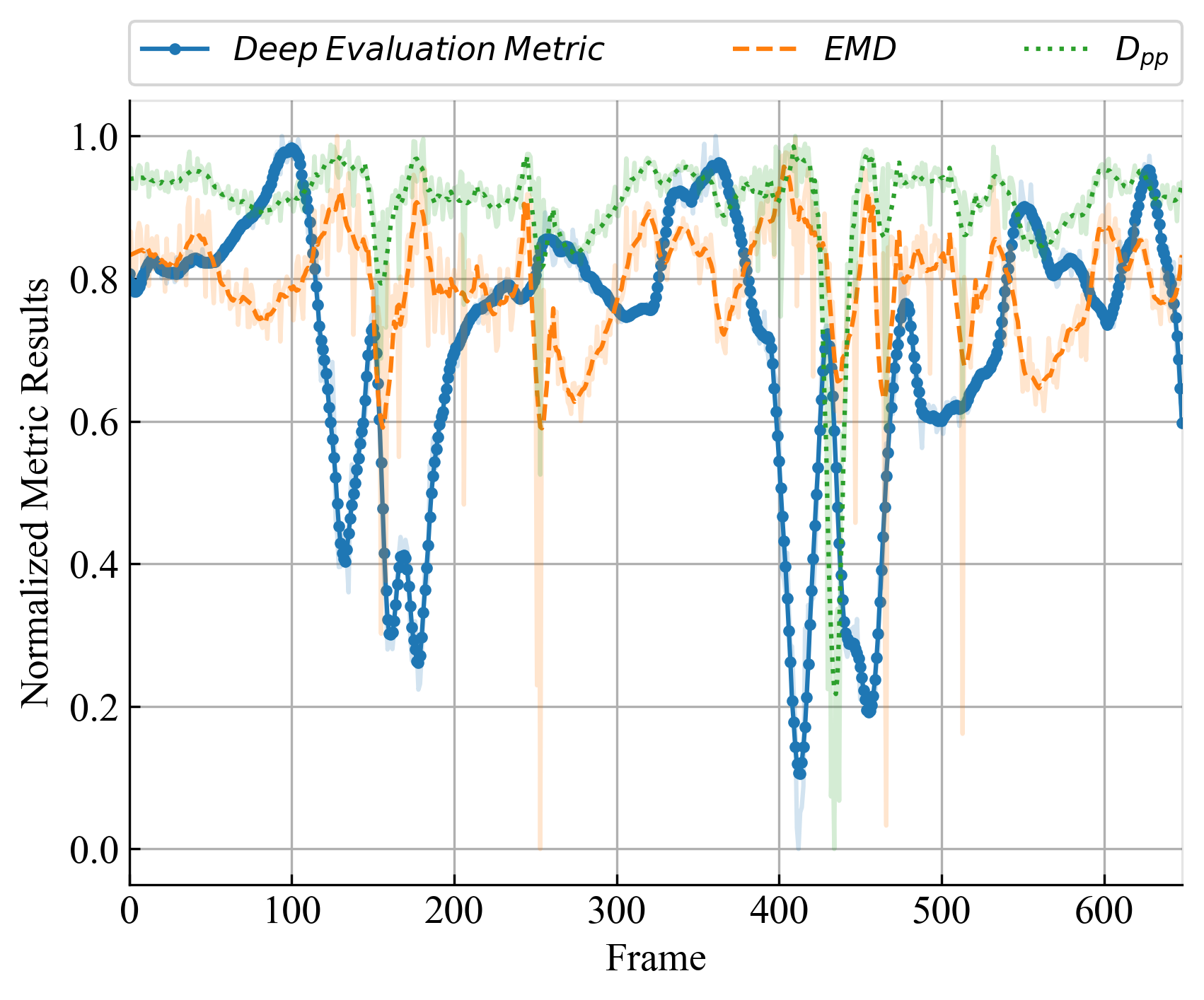}
	}}
	\caption{The solid and moderately transparent lines represent the unfiltered results, while the dashed lines indicate the smoothed point cloud metric results.}
	\label{fig: metrics over frame}
\end{figure}

To further analyze the metrics, the results are plotted over time in Fig. \ref{fig: metrics over frame}. It is particularly apparent that all three metrics indicate a relatively good overall radar model fidelity, particularly EMD ($\mu\!=\!0.79$, $\sigma\!=\!0.09$) and D\textsubscript{pp} ($\mu\!=\!0.90$, $\sigma\!=\!0.09$). However, the proposed Deep Evaluation Metric (DEM) predicts the lowest fidelity with a relatively large standard deviation ($\mu\!=\!0.72$, $\sigma\!=\!0.19$).

While EMD does not indicate any deterioration of the synthetic data in the near ranges (around frame 150 and 430), D\textsubscript{pp} has only a particularly strong minimum peak in the area directly in front of the sensor. This minimum can be reasoned by the strongly increasing number of simulated points, which causes a strong increase in the calculated sum of the individual points. 
Despite that the DEM indicates low sensor model fidelities (minima) in both mentioned near ranges, peaks upwards can be additionally observed in these areas. These are presumably caused by the turning point of the target vehicle, because the object is perpendicular to the sensor in these positions and thus the L-shape disappears, which could result in an abrupt increase. However, in Fig. \ref{fig: radar detections} it is apparent that the number of points differ considerably especially in close range, which should produce a further descent.
The effect that too few simulated points appear at larger distance is not significantly reflected by any of the metrics presented. We assume that the number of points is too small to allow a reliable prediction of the network. With EMD and D\textsubscript{pp}, an insufficient number has apparently no effect on the estimated quality of the simulated point cloud.

\section{Conclusion}
\label{sec: conclusion}
In this paper, we introduced a machine learning-based metric to evaluate the fidelity of synthetically generated radar point clouds. In order to investigate the effectiveness of the proposed method, we used additional conventional metrics and compare their capability to identify characteristic differences between the real and simulated radar data.
We have shown that, in contrast to the conventional metrics used, the proposed deep evaluation method is able to recognize the weaknesses of the synthetic point cloud at close range. However, not all effects such as the insufficient number of points at a long distance could be detected, which none of the metrics succeeded in doing. Overall, the proposed metric shows great potential because it was able to reproduce the intuitive result from a qualitative evaluation much better than the other metrics.

Future work will focus on improving the training data, for example by learning the whole scene perceived by a sensor or including other classes than cars such as pedestrians or cyclists.
Besides the extension of the training data, we will investigate to what degree it is advantageous to include the time information in order to take into account the temporal evolution of objects.

\addtolength{\textheight}{0cm}   


\bibliographystyle{IEEEtran}
\bibliography{MyLibrary}

\end{document}